\documentclass{article}

\usepackage{arxiv}

\usepackage[utf8]{inputenc} % allow utf-8 input
\usepackage[T1]{fontenc}    % use 8-bit T1 fonts
\usepackage{hyperref}       % hyperlinks
\usepackage{url}            % simple URL typesetting
\usepackage{booktabs}       % professional-quality tables
\usepackage{amsfonts}       % blackboard math symbols
\usepackage{nicefrac}       % compact symbols for 1/2, etc.
\usepackage{microtype}      % microtypography
\usepackage{lipsum}
\usepackage{graphicx}
\graphicspath{ {./images/} }

\title{ASCAT: An Arabic Scientific Corpus and Benchmark for Advanced Translation Evaluation}

\author{
Serry Sibaee \\
Prince Sultan University \\
\texttt{serrymrss@gmail.com}
\And
Khloud Al Jallad \\
SySSR \\
\texttt{k.jallad.l@gmail.com}
\And
Zineb Yousfi \\
NAMAA Community \\
\texttt{yousfi.zineb.yz@gmail.com}
\And
Israa Elsayed Elhosiny \\
Independent Linguist \\
\texttt{israaelhosiny@gmail.com}
\And
Yousra El-Ghawi \\
NAMAA Community \\
\texttt{yousraghawi@gmail.com}
\And
Batool Balah \\
NAMAA Community \\
\texttt{batoolnajeh@gmail.com}
\And
Omer Nacar \\
Tuwaiq Academy \\
\texttt{o.najar@tuwaiq.edu.sa}
}

\begin{document}
\maketitle
\begin{abstract}
We present ASCAT (Arabic Scientific Corpus for Advanced Translation), a high-quality English-Arabic parallel benchmark corpus designed for scientific translation evaluation constructed through a systematic multi-engine translation and human validation pipeline. Unlike existing Arabic-English corpora that rely on short sentences or single-domain text, ASCAT targets full scientific abstracts averaging 141.7 words (English) and 111.78 words (Arabic), drawn from five scientific domains: physics, mathematics, computer science, quantum mechanics, and artificial intelligence. Each abstract was translated using three complementary architectures generative AI (Gemini), transformer-based models (Hugging Face \texttt{quickmt-en-ar}), and commercial MT APIs (Google Translate, DeepL) and subsequently validated by domain experts at the lexical, syntactic, and semantic levels. The resulting corpus contains 67,293 English tokens and 60,026 Arabic tokens, with an Arabic vocabulary of 17,604 unique words reflecting the morphological richness of the language. We benchmark three state-of-the-art LLMs on the corpus GPT-4o-mini (BLEU: 37.07), Gemini-3.0-Flash-Preview (BLEU: 30.44), and Qwen3-235B-A22B (BLEU: 23.68) demonstrating its discriminative power as an evaluation benchmark. ASCAT addresses a critical gap in scientific MT resources for Arabic and is designed to support rigorous evaluation of scientific translation quality and training of domain-specific translation models.
\end{abstract}

% keywords can be removed

\section{Introduction}
The rapid growth of scientific literature has intensified the need for reliable domain-specific translation models, particularly for Arabic spoken by over 400 million people yet grossly underrepresented in scientific discourse. This language gap creates a critical accessibility barrier for Arabic-speaking researchers and professionals. A key bottleneck in improving Arabic scientific machine translation (MT) is the scarcity of high-quality parallel corpora that maintain terminological accuracy and conceptual consistency.

This paper presents the construction and analysis of an English-Arabic scientific translation corpus built through a systematic multi-engine translation and human validation process. By combining multiple state-of-the-art translation engines including generative AI (Gemini), transformer-based models (Hugging Face), and commercial MT APIs (Google Translate, DeepL) with rigorous expert validation, our approach balances scalability with quality. The corpus covers full scientific abstracts across diverse domains including physics, mathematics, computer science, quantum mechanics, and artificial intelligence, averaging 141.7 words (English) and 125.4 words (Arabic) per abstract far exceeding the complexity of existing datasets.
Critically, ASCAT is positioned as an evaluation benchmark rather than a large-scale training resource, prioritizing depth of validation over dataset size.
\section{Background}
\label{sec:related}

Existing English-Arabic parallel corpora suffer from several key limitations. General-domain resources such as MultiUN~\cite{eisele-chen-2010-multiun} and OPUS~\cite{tiedemann2012opus} lack the terminological precision required for scientific translation. Domain-specific datasets like DEAST~\cite{deast2026} (33,000 thesis title pairs, $\sim$9 words average) and PEACH~\cite{alsabbagh2024peach} (51,671 healthcare sentence pairs, $\sim$10--12 words average) are too short to capture the syntactic complexity and discourse-level phenomena of scientific abstracts. While Tarjama-25~\cite{hennara2025mutarjim} targets longer sentences ($\sim$75 words), its 5,000-sentence scale is insufficient for training large MT models. ATHAR~\cite{mohammed2025athar} addresses classical Arabic scientific texts, which differ substantially in register from modern scientific writing. Table~\ref{tab:dataset_comparison} provides a structured comparison of these corpora against our dataset. Note that, ASCAT's multi-stage validation makes it benchmark-grade.

\begin{table*}[t]
\centering
\footnotesize
\caption{Comparison of English--Arabic Parallel Corpora}
\label{tab:dataset_comparison}
\begin{tabular}{lccccc}
\hline
\textbf{Dataset} & \textbf{Domain} & \textbf{Size} & \textbf{Type} & \textbf{Avg. Len.} & \textbf{Val.} \\
\hline
DEAST~\cite{deast2026} & Multi-sci. & 33k & Titles & 9 w & Expert \\
PEACH~\cite{alsabbagh2024peach} & Health & 51.7k & Leaflets & 10--12 w & Experts \\
ATHAR~\cite{mohammed2025athar} & Classical & 66k & Historical & -- & Expert \\
Tarjama-25~\cite{hennara2025mutarjim} & Multi-dom. & 5k & Long sent. & 75 w & Prof. \\
\textbf{ASCAT (Ours)} & Multi-sci. & 500 & Abstracts & \textbf{125 / 112 w} & \textbf{Multi-stage} \\
\hline
\end{tabular}
\end{table*}

Multi-engine translation approaches have emerged as a promising strategy, exploiting complementary strengths: statistical MT excels at terminology consistency, neural MT produces fluent output, and large language models handle contextual and idiomatic expressions. Combining their outputs enables comparative analysis and focuses human validation on high-risk segments. Human validation remains essential in scientific translation full post-editing, selective validation guided by confidence scoring, and inter-annotator agreement metrics each play a role in ensuring quality, yet few existing corpora document their validation protocols transparently.

Our dataset addresses these gaps by targeting full scientific abstracts, employing a multi-engine translation pipeline, and applying multi-stage expert validation yielding a resource suitable for both training domain-specific MT models and benchmarking scientific translation quality.

\section{Methodology}
\label{sec:Dataset Construction}

\begin{figure}[h]
    \centering
    \includegraphics[width=1.0\linewidth]{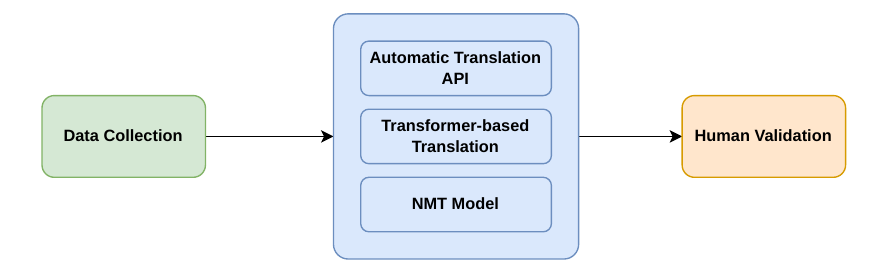}
    \caption{Dataset construction pipeline: Data Collection, Multi-Engine Translation, and Human Validation.}
    \label{fig:pipeline}
\end{figure}

The dataset construction follows a three-stage pipeline Data Collection, Multi-Engine Translation, and Human Validation as illustrated in Figure~\ref{fig:pipeline}.

\subsection{Data Collection}
Scientific abstracts were systematically gathered across multiple domains, including physics, mathematics, computer science, quantum mechanics, and artificial intelligence. To ensure representational diversity and avoid selection bias, abstracts were randomly sampled from papers across these domains, yielding a balanced multi-domain corpus of full-length scientific abstracts.

\subsection{Multi-Engine Translation}
To maximize linguistic diversity and enable comparative analysis, each abstract was translated using three distinct architectures:

\begin{itemize}
    \item \textbf{Generative AI (Gemini):} The Gemini API was used to produce contextually nuanced translations, leveraging its reasoning capabilities to handle complex scientific discourse beyond literal word-for-word mapping.
    \item \textbf{Transformer-based Models (Hugging Face):} The \texttt{quickmt-en-ar} model from the Hugging Face ecosystem was integrated to provide domain-adapted neural translations.
    \item \textbf{Commercial MT APIs (Google Translate \& DeepL):} Industry-standard APIs were employed as high-fluency baselines, enabling systematic comparison against specialized and generative approaches.
\end{itemize}

\subsection{Human Validation}
The final stage involved comprehensive expert review of all translated samples. Validation was conducted by seven domain experts, each holding at minimum a graduate-level degree in either Arabic linguistics or a relevant scientific discipline (physics, mathematics, computer science, or AI). Each validator was assigned abstracts within their area of expertise, ensuring that terminological judgments were made by subject-matter specialists rather than generalist translators. Validators worked independently using a structured checklist, and disagreements were resolved through discussion until consensus was reached.

\begin{table}[ht]
\centering
\footnotesize
\caption{Human Validation Checklist}
\label{tab:validation_checklist}
\begin{tabular}{llc}
\hline
\textbf{Level} & \textbf{Criterion} & \textbf{Binary (Y/N)} \\
\hline
Lexical   & Domain terminology accuracy     & Y \\
Lexical   & Preservation of named entities  & Y \\
Syntactic & Grammatical correctness (Arabic) & Y \\
Syntactic & Sentence structure preservation  & Y \\
Semantic  & Epistemic hedging preserved     & N \\
Discourse & Sentence count consistency      & N \\
\hline
\end{tabular}
\end{table}

Human validators with expertise in both Arabic linguistics and the relevant scientific domains examined translations at the lexical, syntactic, and semantic levels, correcting terminology errors, structural inconsistencies, and meaning deviations. A detailed analysis of the most frequent translation errors identified during this phase is provided in section \ref{data-anls}.

\section{Dataset Analysis}
\label{data-anls}

\subsection{Domain Distribution}
The dataset was strategically curated to ensure broad coverage of modern scientific disciplines. The corpus spans six primary domains: Quantum Mechanics, Artificial Intelligence, Computer Science, Mathematics, Physics. This multi-domain design ensures that the corpus captures the terminological diversity and stylistic variation inherent to different scientific fields, making it suitable for training and evaluating general-purpose scientific MT systems rather than narrow single-domain models. The deliberate balance across domains also mitigates the risk of domain-specific lexical bias, where a model trained predominantly on one field may underperform on others.

\subsection{Sentence Length Analysis}
To assess the linguistic complexity of the corpus, we analyzed word and character distributions across all abstract pairs. Table~\ref{tab:length_stats} summarizes the key statistics for both the English source and Arabic target abstracts.

\begin{table}[ht]
\centering
\footnotesize
\caption{Sentence Length Statistics}
\label{tab:length_stats}
\begin{tabular}{lcccccc}
\hline
\textbf{Lang} & \textbf{Min} & \textbf{Med} & \textbf{Mean} & \textbf{Max} & \textbf{SD} & \textbf{Mean Ch.} \\
\hline
EN & 3 & 113 & 125.3 & 297 & 64.0 & 822 \\
AR & 4 & 100 & 111.8 & 315 & 58.9 & 695 \\
\hline
\end{tabular}
\end{table}

The English abstracts average 125.31 words per abstract (median: 113), with a standard deviation of 63.99 words, indicating substantial length variability across domains and paper types. Arabic translations average 111.78 words (median: 100) with a standard deviation of 58.87. The consistently lower word count in Arabic relative to English is linguistically expected, as Arabic's rich morphological system and agglutinative properties allow it to encode more information per word through affixation and cliticization.

The high standard deviation in both languages reflects the natural heterogeneity of scientific abstracts: concise theoretical results may be presented in fewer than 50 words, while comprehensive experimental studies may span close to 300 words. This variability is a desirable property for a training corpus, as it exposes MT models to a wide spectrum of abstract lengths and structural complexities.

\subsection{Vocabulary and Lexical Richness}
Table~\ref{tab:vocab_stats} presents the vocabulary and token-level statistics for both language sides of the corpus.

\begin{table}[ht]
\centering
\small
\caption{Vocabulary Statistics}
\label{tab:vocab_stats}
\begin{tabular}{lcccc}
\hline
\textbf{Lang} & \textbf{Tokens} & \textbf{Unique words} & \textbf{TTR} & \textbf{Mean Sent.} \\
\hline
EN & 67,293 & 12,685 & 0.19 & 7.16 \\
AR & 60,026 & 17,604 & 0.29 & 6.99 \\
\hline
\end{tabular}
\end{table}

Several notable observations emerge from the vocabulary analysis. First, despite having fewer total tokens (60,026 vs.\ 67,293), the Arabic side exhibits a substantially larger vocabulary size (17,604 unique words vs.\ 12,685 in English). This is reflected in the Type-Token Ratio (TTR), a standard measure of lexical diversity, where Arabic scores 0.2933 compared to English's 0.1885. The higher Arabic TTR is consistent with the morphological complexity of Arabic, where a single root can generate dozens of surface forms through derivational and inflectional processes, resulting in a much larger effective vocabulary space. This has direct implications for MT model design, as Arabic-side models require larger vocabularies or subword tokenization strategies (e.g., BPE or SentencePiece) to adequately cover the target language's lexical space.

Second, both language sides exhibit comparable mean sentence counts per abstract (7.16 for English and 6.99 for Arabic), confirming that the translation process preserved the discourse segmentation of the original abstracts without undue merging or splitting of sentences. This structural fidelity is important for training models that must learn to produce coherent, multi-sentence scientific discourse rather than isolated sentence translations.

% Together, these statistics confirm that our corpus occupies a distinct and underserved niche in the landscape of English-Arabic parallel corpora: long-form, lexically rich, domain-diverse scientific abstracts that challenge MT systems at the discourse, syntactic, and terminological levels simultaneously.

\section{Evaluation}
\label{sec:evaluation}

The following evaluation demonstrates ASCAT's utility as a discriminative benchmark — a role for which validation quality and domain complexity matter more than corpus size.

To assess the quality of our corpus and benchmark the performance of state-of-the-art translation systems on scientific English-Arabic translation, we evaluated three large language models against our human-validated reference translations. The models evaluated are \textbf{GPT-4o-mini} (OpenAI), \textbf{Gemini-3.0-Flash-Preview} (Google DeepMind), and \textbf{Qwen3-235B-A22B} (Alibaba Cloud). Each model was prompted to translate the English source abstracts, and the outputs were scored against the human-validated Arabic references using two standard automatic evaluation metrics: BLEU and ROUGE.

% \subsection{Evaluation Metrics}

% \textbf{BLEU} (Bilingual Evaluation Understudy)~\cite{papineni2002bleu} measures n-gram precision between the hypothesis and reference translation, penalizing overly short outputs via a brevity penalty. It remains the most widely adopted metric for MT evaluation despite known limitations in morphologically rich languages such as Arabic.

% \textbf{ROUGE} (Recall-Oriented Understudy for Gisting Evaluation)~\cite{lin2004rouge} measures the overlap between hypothesis and reference at the unigram (ROUGE-1), bigram (ROUGE-2), and longest common subsequence (ROUGE-L) levels, emphasizing recall alongside precision. ROUGE-L is particularly informative for long-form scientific text, as it captures structural similarity beyond local n-gram overlap.

\subsection{Results}

Table~\ref{tab:evaluation} presents the automatic evaluation results for all three models.

\begin{table}[ht]
\centering
\footnotesize
\caption{Automatic Evaluation on EN–AR Scientific Translation}
\label{tab:evaluation}
\begin{tabular}{lcccc}
\hline
\textbf{Model} & \textbf{BLEU} $\uparrow$ & \textbf{R-1} $\uparrow$ & \textbf{R-2} $\uparrow$ & \textbf{R-L} $\uparrow$ \\
\hline
GPT-4o-mini & \textbf{37.07} & \textbf{0.590} & \textbf{0.476} & \textbf{0.586} \\
Gem-3.0-Flash & 30.44 & 0.530 & 0.390 & 0.522 \\
Qwen3-235B & 23.68 & 0.537 & 0.410 & 0.531 \\
\hline
\end{tabular}
\end{table}

\subsection{Discussion}

GPT-4o-mini achieves the highest scores across all metrics (BLEU: 37.07, ROUGE-L: 0.5862), demonstrating strong alignment with human-validated references at both local and discourse levels a notable result for a compact, efficiency-oriented model.

Gemini-3.0-Flash-Preview scores moderately (BLEU: 30.44, ROUGE-L: 0.5219), with a notable gap between its BLEU and ROUGE-1 scores suggesting adequate content coverage but less precise n-gram sequence matching, likely due to greater paraphrastic variation in its outputs.

Qwen3-235B-A22B, despite being the largest model by parameter count, achieves the lowest BLEU (23.68) while remaining competitive on ROUGE-1 (0.5369) and ROUGE-2 (0.4099). This divergence suggests lexically relevant but structurally distant translations relative to the reference, possibly reflecting differences in Arabic stylistic tendencies within its training data.

The performance gap of up to 13.4 BLEU points between systems confirms the discriminative power of our corpus as an evaluation benchmark. The moderate absolute scores across all models even for the best-performing GPT-4o-mini reflect the inherent difficulty of long-form scientific Arabic translation, reinforcing the need for domain-specific resources such as the corpus introduced in this work.

\section{Limitations and Future Work}
\label{sec:limitations}

Despite the quality of our corpus, several limitations warrant acknowledgment. First, the dataset's size (500 abstracts) is by design, as benchmark quality requires intensive human validation that doesn't scale arbitrarily, prioritizing depth over breadth. Second, the dataset remains unevenly distributed across scientific disciplines, which may affect the generalizability of trained models to underrepresented fields. Third, evaluation relied primarily on automatic metrics (BLEU, ROUGE), which do not fully capture qualitative dimensions such as semantic adequacy, epistemic nuance, and terminological felicity.

Scientific translation also presents inherent linguistic challenges that current MT systems struggle to handle reliably. These include interdisciplinary terminological ambiguity (e.g., \textit{state} in physics vs.\ mathematics), non-standardized technical terms (e.g., \textit{dimensional crossover}), preservation of epistemic hedging devices (\textit{may suggest}, \textit{appears to}), and the handling of acronyms and eponymous constructs (e.g., \textit{Wasserstein distance}, \textit{Green's tensor}) where no standardized Arabic equivalent exists.

Future work should focus on three directions: (1) expanding the corpus toward a more balanced domain distribution, (2) incorporating large-scale human evaluation to complement automatic metrics, and (3) fine-tuning domain-adapted models on the corpus to assess downstream translation improvements. These steps would move the field closer to MT systems capable of genuinely cross-disciplinary scientific translation.

\section{Conclusion}
\label{sec:conclusion}

This paper presented ASCAT, a high-quality English-Arabic parallel benchmark corpus designed for scientific translation evaluation constructed through a systematic multi-engine translation and expert validation pipeline. By combining generative AI (Gemini), transformer-based models (Hugging Face), and commercial MT APIs (Google Translate, DeepL) with rigorous human validation, ASCAT addresses a critical gap in Arabic scientific MT resources offering abstract-level linguistic complexity, multi-domain coverage, and methodological transparency absent from existing corpora.

Corpus analysis confirmed the richness of the dataset: Arabic's higher Type-Token Ratio (0.29 vs.\ 0.19) and larger vocabulary (17,604 vs.\ 12,685 unique words) highlight the morphological challenges that MT systems must overcome. Benchmarking three state-of-the-art LLMs revealed a performance gap of up to 13.4 BLEU points between systems, demonstrating ASCAT's discriminative power as an evaluation benchmark while underscoring the persistent difficulty of long-form scientific Arabic translation.

We believe ASCAT represents a meaningful step toward closing the language gap in scientific communication for Arabic speakers, and we hope it serves as a foundation for future advances in domain-specific Arabic MT research.

\bibliographystyle{unsrt}  
\bibliography{references}  %%% Remove comment to use the external .bib file (using bibtex).

@inproceedings{eisele-chen-2010-multiun,
    title = "{M}ulti{UN}: A Multilingual Corpus from United Nation Documents",
    author = "Eisele, Andreas  and
      Chen, Yu",
    editor = "Calzolari, Nicoletta  and
      Choukri, Khalid  and
      Maegaard, Bente  and
      Mariani, Joseph  and
      Odijk, Jan  and
      Piperidis, Stelios  and
      Rosner, Mike  and
      Tapias, Daniel",
    booktitle = "Proceedings of the Seventh International Conference on Language Resources and Evaluation ({LREC}'10)",
    month = may,
    year = "2010",
    address = "Valletta, Malta",
    publisher = "European Language Resources Association (ELRA)",
    url = "https://aclanthology.org/L10-1473/"
}

@inproceedings{tiedemann2012opus,
  title={Parallel Data, Tools and Interfaces in OPUS},
  author={Tiedemann, J{\"o}rg},
  booktitle={Proceedings of LREC 2012},
  pages={2214--2218},
  year={2012}
}

@article{deast2026,
  title={DEAST: A dataset for English-Arabic scientific translation and vice versa},
  author={Author, M. A. and Others},
  journal={Data in Brief},
  volume={64},
  pages={112381},
  year={2026},
  doi={10.1016/j.dib.2025.112381}
}

@article{alsabbagh2024peach,
  title={PEACH: A sentence-aligned Parallel English-Arabic Corpus for Healthcare},
  author={Al-Sabbagh, Rania},
  journal={Corpora},
  volume={19},
  number={3},
  pages={395--410},
  year={2024},
  doi={10.3366/cor.2024.0320}
}

@inproceedings{mohammed2025athar,
  title={ATHAR: A High-Quality and Diverse Dataset for Classical Arabic to English Translation},
  author={Mohammed, Mahmoud S. and Khalil, Mohamed},
  booktitle={Proceedings of the Third Arabic Natural Language Processing Conference},
  pages={97--106},
  year={2025}
}

@article{hennara2025mutarjim,
  title={Mutarjim: Advancing Bidirectional Arabic-English Translation with a Small Language Model},
  author={Hennara, K. and Hreden, M. and Hamed, M. M. and Aldallal, Z. and Chrouf, S. and AlModhayan, S.},
  journal={arXiv preprint arXiv:2505.17894},
  year={2025}
}

%/begin{thebibliography}{1}
%\end{thebibliography}

\end{document}